\documentclass[tightenlines, twocolumn, notitlepage, nofootinbib, superscriptaddress]{revtex4}

\usepackage{algorithm, algorithmic}
\usepackage{amsmath}
\usepackage{amssymb}
\usepackage{bibunits}
\usepackage{array}
\usepackage{enumerate}
\usepackage{booktabs}
\usepackage[version=4]{mhchem}
\usepackage[dvipsnames]{xcolor}
\usepackage{graphicx}
\usepackage{multirow}
\usepackage{natbib}
\usepackage{tabularx}
\usepackage{units}
\usepackage{hyperref}
\hypersetup{
    colorlinks = true,
    linkbordercolor = {white},
    anchorcolor     = {black},
    citecolor       = NavyBlue,
    filecolor       = {cyan},
    menucolor       = {red},
    runcolor        = {cyan},
    urlcolor        = OliveGreen
}

\usepackage{float}
\newfloat{Reaction}{hbtp}{tbp}

\usepackage{tikz}
\usetikzlibrary{calc}
\usetikzlibrary{positioning}
\tikzstyle{my_box} = [black, very thick, rounded corners, align = center, anchor = north west]

\graphicspath{{./}}
\setcitestyle{super}

\AtBeginDocument{
\heavyrulewidth=.08em
\lightrulewidth=.05em
\cmidrulewidth=.03em
\belowrulesep=.65ex
\belowbottomsep=0pt
\aboverulesep=.4ex
\abovetopsep=0pt
\cmidrulesep=\doublerulesep
\cmidrulekern=.5em
\defaultaddspace=.5em
}

\newcommand{\error}[2]{$#1$ {\footnotesize $\pm\,#2$}}
\newcommand{\smerr}[2]{${\bf #1}$ {\footnotesize $\pm\,#2$}}
\newcommand{\vect}[1]{\boldsymbol{#1}}

\newcommand{\name}{PHOENICS }
\newcommand{\namecomma}{PHOENICS, }

\begin{document}


	\title{\Large{}PHOENICS: A universal deep Bayesian optimizer}

	\date{\today}

	\author{Florian H\"ase}
	\author{Lo\"ic M. Roch}
	\author{Christoph Kreisbeck}
	\affiliation{Department of Chemistry and Chemical Biology, Harvard University, Cambridge, Massachusetts, 02138, USA}
	\author{Al\'an Aspuru-Guzik}
	\email{alan@aspuru.com}
	\affiliation{Department of Chemistry and Chemical Biology, Harvard University, Cambridge, Massachusetts, 02138, USA}
	\affiliation{Senior Fellow, Canadian Institute for Advanced Research, Toronto, Ontario M5G 1Z8, Canada}


	\begin{abstract}
		In this work we introduce \namecomma a probabilistic global optimization algorithm combining ideas from Bayesian optimization with concepts from Bayesian kernel density estimation. We propose an inexpensive acquisition function balancing the explorative and exploitative behavior of the algorithm. This acquisition function enables intuitive sampling strategies for an efficient parallel search of global minima. 
		The performance of \name is assessed via an exhaustive benchmark study on a set of 15 discrete, quasi-discrete and continuous multidimensional functions. Unlike optimization methods based on Gaussian processes (GP) and random forests (RF), we show that \name is less sensitive to the nature of the co-domain, and outperforms GP and RF optimizations. We illustrate the performance of \name on the Oregonator, a difficult case-study describing a complex chemical reaction network. We demonstrate that only \name was able to reproduce qualitatively and quantitatively the target dynamic behavior of this nonlinear reaction dynamics. We recommend \name for rapid optimization of scalar, possibly non-convex, black-box unknown objective functions.
	\end{abstract}

	\maketitle

	\begin{bibunit}[unsrt]
		
	\section{Introduction}

	Optimization problems are ubiquitous in a rich variety of disciplines ranging from science to engineering and can take various facets: finding the lowest energy state of a system, searching for the best set of parameters to improve industrial processes, or identifying the best business strategies to maximize profit. They also have a rich history in chemistry. For example, conditions for chemical reactions are optimized with systematic methods like design of experiments (DOE).\cite{Fisher1937, Box2005, Anderson2016} Only recently, optimization procedures assisted chemists in finding chemical derivatives of given molecules to best treat a given disease,\cite{Negoescu2011} finding candidates for organic photovoltaics,\cite{Lopez2017} predicting reaction paths,\cite{Wei2016, Coley2017, Segler2017} adjusting robotics control parameters for improving gait speed and smoothness,\cite{Lizotte2007} or in automated experimentation.\cite{Nikolaev2014, Nikolaev2016, Cronin2017}

	Theoretically, optimization problems are formulated with an objective function, which for a given set of parameters returns a value representing the cost associated with this set of parameters. In the past, a variety of algorithms have been developed. For example, gradient based algorithms are efficient at finding local minima, but are not suited for global optimization of non-convex objective functions. 

	Lately, the development of methods for finding the global optimum of non-convex, expensive to evaluate objective functions has gained resurgence as a very active field of research. Simplistic approaches may consist of random searches, or systematic grid searches. While random searches have often been shown to be suitable in the context of hyperparameter optimization for machine learning models,\cite{Bergstra2011, Bergstra2012} systematic grid search approaches like DOE were successfully applied to real-life experimentation planning.\cite{Fisher1937, Box2005, Anderson2016} More sophisticated methods built on Bayesian optimization approaches have emerged as a popular and efficient alternative during the last decade.\cite{Kushner1964, Mockus1975, Mockus1982, Snoek2012, Srivinas2012, smac2017}

	The typical procedure of Bayesian optimization schemes consists of two major steps: First, find an approximation to the landscape of the objective function; and Second, propose the next point to be evaluated based on this approximation. Several different machine learning models have been suggested for approximating the objective function landscape, ranging from random forests (RFs),\cite{Hutter2011, Hutter2012, smac2017} over Gaussian processes (GPs),\cite{Snoek2012, Snoek2014} to Bayesian neural networks (BNNs).\cite{Snoek2015, Springenberg2016} Likewise, a variety of methods for proposing new parameter points from probabilistic models is frequently used.\cite{Mockus1975, Jones1998, Snoek2012, Hernandez2014, Hernandez2015, Srivinas2010}

	Although Bayesian optimization has been successfully employed for a variety of applications,\cite{Gomez2016, Pyzer2016, Hase2017, Martinez2017, Ju2017, Cornejo2017, Kikuchi2017} we identify three limitations:

	
	\begin{enumerate}[(i)]
		\labelsep = 1.2em 
		\itemsep  = -3pt
		\item GPs typically perform well on objective functions with continuous co-domain, but are outperformed by RFs on objective functions with discrete and quasi-discrete co-domain and categorical parameters. For an unknown objective function it would be most desirable to have a method, which performs well on both cases.
		\item Traditionally, Bayesian optimization methods are sequential in nature, which prevents parallel evaluations of the objective function. For parallel evaluation, however, an acquisition function enabling the generation of multiple informative parameter points is needed. Despite prior works\cite{Snoek2012, Wang2017, Contal2013, Desautels2014} this remains an open challenge.
		\item Bayesian optimization is only applied efficiently if the computational cost of one optimization iteration is lower than the cost of one objective function evaluation. BNNs have been introduced to circumvent the cubical scaling of GPs in the Bayesian inference step. However, acquisition function evaluations are computationally more expensive for BNNs than for GPs. The computational cost of Bayesian optimization typically limits its applicability to low-dimensional parameter spaces for which the optimum can be found in relatively few objective function evaluations. 
	\end{enumerate}

	The Probabilistic Harvard Optimizer Exploring Non-Intuitive Complex Surfaces (PHOENICS) algorithm introduced in this study addresses the aforementioned limitations by supplementing ideas from Bayesian optimization with concepts from Bayesian kernel density estimation. Specifically, we use BNNs to estimate kernel distributions associated with a particular objective function value from observed parameter points. Our approach differs from the traditional use of BNNs in the Bayesian optimization context, where objective function values are predicted from BNNs directly. Employing the estimated kernel distributions, we can construct a simple functional form of the approximation to the objective function. As a consequence, the computational cost of \name scales linearly with the dimensionality and the number of observations, without the cost of a full Bayesian evaluation of the BNN. \name is available for download on GitHub.\cite{githubRepo}

	We propose an inexpensive acquisition function, which enables intuitive strategies for efficient parallelization. This is achieved by simultaneously proposing multiple parameter points with different sampling policies at negligible additional cost. Those policies are biased towards exploration or exploitation by tuning an intuitive hyperparameter, balancing the explorative and exploitative behavior of the algorithm. A synergistic effect is observed when proposing batches of parameter points with different sampling policies. Our batching policy not only helps to accelerate the optimization process, but also reduces the total number of required function evaluations. It is therefore to be seen as an improvement over trivial parallelization.  

	We provide an exhaustive benchmark study of \name and compare it to well-established Bayesian optimization methods based on GPs and RFs. We demonstrate that \name performs equally well on analytic benchmark functions with discrete and continuous co-domains. Further we apply \name to the Oregonator, an oscillating chemical model system governed by a set of non-linear coupled differential equations. On this difficult optimization problem, we show that only \name reproduces the target dynamic behavior.

	In what follows, we start with a brief overview of related works in Sec.~\ref{sec:background}. Then we detail the mathematical formulation of \name in Sec.~\ref{sec:methods}. In Sec.~\ref{sec:results} we discuss performance results of \name on analytic benchmark functions and compare to other Bayesian optimization methods. Before concluding we further demonstrate the strength of our approach on the Oregonator.


	\section{Background and related work}\label{sec:background}

		Bayesian optimization is a gradient-free, sequential strategy for the global optimization of possibly noisy black-box functions, which we denote with $f$ from hereon.\cite{Kushner1964, Mockus1975, Mockus1982, Osborne2009} Bayesian optimization aims to find the global optimum of a given objective function $f$ within as few evaluations as possible. It consists of two major steps: (i) constructing a probabilistic approximation to $f$ and (ii) proposing new parameter points for querying $f$ based on the probabilistic approximation.

		The probabilistic approximation is constructed by first conditioning $f$ on a prior $\phi_\text{prior}(\vect{\theta})$ over the functional form, which is described by parameters $\vect{\theta}$. The parameters $\vect{\theta}$ of the prior distribution are refined based on observations of $n$ pairs $\mathcal{D}_n$ of parameter values $\vect{x}_k$ and corresponding objective function values $f_k = f(\vect{x}_k)$, as described by Eq.~\ref{eq:obs}

		\begin{align}\label{eq:obs}
			\mathcal{D}_n = \big\{\big(\vect{x}_k, f_k\big)\big\}_{k=1}^n.
		\end{align}

		The functional prior distribution $\phi_\text{prior}$ is updated based on observations $\mathcal{D}_n$ using \emph{Bayes' theorem} (see Eq.~\ref{eq:bayes_theorem}) to obtain the posterior distribution $\phi_\text{post}$

		\begin{align}\label{eq:bayes_theorem}
			\phi_\text{post}(\vect{\theta} | \mathcal{D}_{n}) = \frac{p(\mathcal{D}_{n} | \vect{\theta}) \enskip \phi_\text{prior}(\vect{\theta})}{p(\mathcal{D}_{n})}.
		\end{align}

		With more and more observations $\mathcal{D}_n$ the posterior distribution $\phi_\text{post}$ yields a better approximation and eventually converges to the objective function in the limit of infinitely many distinct observations. 

		In the second step of the general Bayesian optimization procedure, this probabilistic approximation to $f$ is used to propose points in parameter space at which $f$ is to be evaluated next. Bayesian optimization is efficient in cases where evaluations of the constructed approximation are much cheaper than evaluations of $f$. Under this assumption, the constructed approximation is inexpensively queried to reason about regions in parameter space where the global optimum could be located. Decisions about parameter points at which $f$ should be evaluated are typically chosen from a so-called acquisition function constructed from the probabilistic approximation. Bayesian optimization therefore relies on both an accurate approximation to the objective function and also the formulation of an efficient acquisition function.

		\subsection{Constructing the objective function approximation}

			A popular choice for modeling the functional prior $\phi_\text{prior}$ on the objective function are Gaussian processes (GPs).\cite{Martinez2009, Osborne2009, Snoek2012, Desautels2014} GPs associate every point in the parameter domain with a normally distributed random variable. These normal distributions are then constructed via a similarity measure between observations given by a kernel function. A GP therefore provides a flexible way of finding analytic approximations to the objective function, from which uncertainty information can be obtained. Training a GP via Bayesian inference, however, is computationally costly as it involves the inversion of a dense covariance matrix. The construction of a GP therefore scales cubically in the dimensionality of the parameter space and with the number of observations. Due to this limitation, GPs are typically used in relatively low dimensional problems with an optimum that can be found in relatively few objective function evaluations. 

			Another popular choice are random forests (RFs).\cite{Breiman2001, Hutter2011, Hutter2012, smac2017} RFs are a collection of regression trees, which, in contrast to decision trees, have real numbers at their leaves. RFs have been shown to perform particularly well for categorical input data and classification tasks. RFs are therefore successfully applied to objective functions with discrete or quasi-discrete co-domain. The computational cost of training a RF scales as $\mathcal{O}(n \log n)$ with the number of observations and linearly with the dimensionality of the parameter space. Although RFs can be trained rapidly, model uncertainty needs to be estimated empirically. 

			Recently, Bayesian neural networks (BNNs) have been employed in Bayesian optimization,\cite{Snoek2015, Springenberg2016} retaining the flexibility and well-calibrated uncertainty of GPs but at a computational scaling comparable to RFs. In contrast to traditional neural networks, weights and biases for neurons in BNNs are not just single numbers but instead sampled from a distribution. BNNs are trained by updating the distributions from which weights and biases are sampled according to Eq.~\ref{eq:bayes_theorem}. Infinitely large BNNs were proven to behave like GPs.\cite{Williams1997}

		\subsection{Acquisition functions}

			Once a probabilistic model approximating the objective function has been constructed, new predictions about the location of the global optimum are made based on an acquisition function. The ideal acquisition function finds the adequate balance between exploration and exploitation. Exploration of the entire parameter space should be favored when no observations in vicinity to the global optimum have been made yet and the acquisition function should only sample close to the global optimum once its general location has been determined.

			One of the earliest and most widely applied acquisition functions is \emph{expected improvement} and variants thereof.\cite{Mockus1975, Jones1998, Snoek2012} Expected improvement aims to measure the expected amount by which an observation of a point in parameter space improves over the current best value. Exploration and exploitation are implicitly balanced based on the posterior mean and the estimated uncertainty.

			More recently, alternative formulations of acquisition functions have been developed. The \emph{upper confidence bound} method exploits confidence bounds for constructing an acquisition function which minimizes regret.\cite{Srivinas2010} Variants of this acquisition function have been designed specifically to be applied in higher dimensional parameter spaces.\cite{Contal2013, Desautels2014} \emph{Predictive entropy} estimates the negative differential entropy of the location of the global optimum given the observations.\cite{Hernandez2014, Hernandez2015}

		\subsection{Batched Bayesian optimization}

			Bayesian optimization traditionally relies on a sequential exploration strategy. For every parameter point suggested from the Bayesian optimizer, the objective function is evaluated to increase the number of observations and refine the probabilistic approximation. In many real-life applications, however, the evaluation of the objective function is time consuming and could benefit from parallelization. 

			A single acquisition function, which proposes only a single new parameter point for evaluation, is not sufficient for batched objective function evaluations. This issue has been addressed in several recent studies, which employ a diversity criterion when proposing a new batch of parameter points to enhance exploration.\cite{Contal2013, Kathuria2016, Wang2017} Applying this diversity criterion, however, becomes computationally expensive for larger batches.


	\section{Methods and computational details}\label{sec:methods}

	In this section we present the mathematical formulation of \namecomma our new algorithm designed for an efficient optimization of scalar, possibly non-convex, black-box objective functions $f$ on the compact subset $\mathcal{X} \in \mathbb{R}^d$. \name focuses on the problem of finding the global minimum

	\begin{align}\label{eq:goal_bopt}
		\vect{x}^* = \underset{\vect{x} \in \mathcal{X}}{\text{argmin}} \, f(\vect{x}).
	\end{align}

	We assume that evaluations of the objective function $f$ are expensive, where the cost could be related to any budgeted resource such as required run-time, experimental synthesis of chemical compounds, computing resources and others.

\begin{figure}[!ht!]
        	\begin{tikzpicture}[scale = 1.0, on grid]

		\begin{scope}
			\node (start) [draw, my_box, text width = 4.1cm, fill = green!50!black!30] at (0, 0) {Starting optimization};
		\end{scope}

		\begin{scope}[yshift = -4 em]
			\node (A) [draw, my_box, text width = 4.1cm, fill = black!15] at (0, 0) {{\LARGE \phantom{+}} \\ training model on current observations (Sec.~\ref{sec:approximation})};
			\node     [draw, my_box, text width = 4.1cm, fill = blue!30] at (0, 0) {Constructing approximation};
		\end{scope}

		\begin{scope}[yshift = -4 em, xshift = 14 em]
			\node (B) [draw, my_box, text width = 3.8cm, fill = black!15] at (0, 0) {{\LARGE \phantom{+}} \\ balancing exploration / exploitation (Sec.~\ref{sec:acquisition_function})};
			\node     [draw, my_box, text width = 3.8cm, fill = blue!30] at (0, 0) {Constructing acquisition};
		\end{scope}

		\begin{scope}[yshift = -11 em, xshift = 14 em]
			\node (C) [draw, my_box, text width = 3.8cm, fill = black!15] at (0, 0) {{\LARGE \phantom{+}} \\ proposing new parameter points (Sec.~\ref{sec:proxy_optimization})};
			\node     [draw, my_box, text width = 3.8cm, fill = blue!30] at (0, 0) {Optimizing acquisition};
		\end{scope}

		\begin{scope}[yshift = -11 em]
			\node (D) [draw, my_box, text width = 4.1cm, fill = black!15] at (0, 0) {{\LARGE \phantom{+}} \\ running the costly \\ experiment / simulation};
			\node     [draw, my_box, text width = 4.1cm, fill = orange!30] at (0, 0) {Evaluating objective};
		\end{scope}

		\draw[->, very thick] ($(start.south) - (1.0, 0)$) -- ($(A.north) - (1.0, 0)$);
		\draw[->, very thick] ($(A.north) + (0.5, 0)$) |- (5, -0.8) -| ($(B.north) - (0.5, 0)$);
		\draw[->, very thick] ($(B.south) - (0.5, 0)$) -- ($(C.north) - (0.5, 0)$);
		\draw[->, very thick] ($(C.south) - (0.5, 0)$) |- (5, -5.5) -| ($(D.south) + (0.5, 0)$);
		\draw[->, very thick] ($(D.north) + (0.5, 0)$) -- ($(A.south) + (0.5, 0)$);

	\end{tikzpicture}
        \caption{General workflow of the optimization procedure introduced in this study.}
        \label{fig:workflow}
    \end{figure}
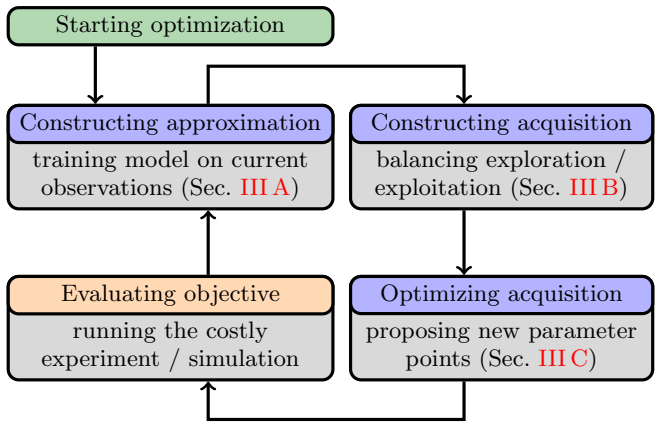

	The overall workflow of \name is schematically represented in Fig.~\ref{fig:workflow} and follows the general principles of traditional Bayesian optimization methods outlined in Sec.~\ref{sec:background}.

	\name supplements ideas from Bayesian optimization with concepts from Bayesian kernel density estimation (BKDE).\cite{Escobar1995} The former relies on the construction of an approximation to the objective function, which is faster to evaluate than the objective function, and, thus, can be used for proxy optimization. BKDE corresponds to a generalization of kernel density estimation and provides a probabilistic model in which the kernel is constantly evolving to best approximate the density of observed parameter points using Bayesian inference.

	Inspired by BKDE, we use BNNs to estimate kernel distributions associated with a particular objective function value from observed parameter points. It has already been reported in the literature\cite{Snoek2015, Springenberg2016} that BNNs are flexible objective function approximators, and have a favorable linear scaling with regard to the number of observations. However, the evaluation of a BNN in a fully Bayesian treatment comes with a significant computational cost. Consequently, the formulation of a much simpler probabilistic model with similar flexibility would be most desirable. 

	\name provides the adequate balance between flexibility and favorable scaling. Unlike traditional approaches, we suggest not to use BNNs for approximating the objective function directly. Instead, \name uses BNNs to approximate the kernel distribution, which yield a particular objective function value and construct the approximation from the averaged densities of all observations. 

	In this procedure, we only need to construct the kernel densities from the BNN once to then formulate a much simpler functional form of the objective function approximation. This allows for much faster evaluation of a given parameter point $\vect{x}$ without further evaluations of the BNN. Details are presented in Sec.~\ref{sec:approximation}. The procedure maintains the flexibility of a BNN at a much lower evaluation cost. As a matter of fact, the simple functional form of the approximation to the objective function provided by \name benefits from the linear scaling with the number of observations, avoiding, however, the cost of a repetitive full Bayesian evaluation of the BNN.


	\subsection{Approximating the objective function}\label{sec:approximation}


		We suggest to train BNNs to estimate the parameter kernel density from the observed parameter points in an autoencoder-like architecture. A particular realization of the BNN represents a map projecting parameter points into the parameter space, i.e $\text{BNN}: \mathbb{R}^d \rightarrow \mathbb{R}^d$. Thereby, we can construct an estimate to the parameter kernel density, which corresponds to a particular observed objective function value. 

		The BNN architecture for all benchmarks in this study consisted in three layers. The dimensionality of the input layer was given by the dimensionality of the parameter space $k$. We chose to model all hidden layers with $50$ units. As described by Eqs.~\ref{eq:network_0} to \ref{eq:network_2}, all layers but the last were connected by hyperbolic tangents; the last layer using a sigmoid activation function

		\begin{align}
			\phi_1          &=    \tanh(x \cdot w_0 + b_0), \label{eq:network_0}\\
			\phi_2          &=    \tanh(\phi_1 \cdot w_1 + b_1), \label{eq:network_1} \\
			\phi_3          &=    \text{sigmoid}(\phi_2 \cdot w_2 + b_2), \label{eq:network_2} \\
			\phi_\text{out} &\sim \mathcal{N}(\phi_3, \tau_n). \label{eq:network_3}
		\end{align}

		Priors for weights and biases of the BNN architecture were chosen to be normal distributions with zero mean $\mu_i = 0$ and unit standard deviation $\sigma_i = 1$ 

		\begin{align}
			w_i &\sim \mathcal{N}(\mu_i, \sigma_i), \\
			b_i &\sim \mathcal{N}(\mu_i, \sigma_i).
		\end{align}

		The output layer $\phi_3$ is used to predict the distributions of means of Gaussian distributions with precisions $\tau_n$, which depend on the number of observations $n$ (see Eq.~\ref{eq:network_3}). The precision $\tau_n$ of these Gaussian distributions are sampled from a Gamma distribution $\tau_n \sim \Gamma(\alpha, \beta)$ with prior hyperparameters $\alpha = 12 n^2$ and $\beta = 1$, where $n$ is the number of observations. With this choice of parameters the precision of the Gaussian distribution increases with the number of observations. Details on the particular choice are provided in the supplementary information (see Sec.~\ref{sec:supp_shrinkage}). Note that the parameter space is rescaled to the unit hypercube prior to training the model.

		The BNN is trained via Bayesian inference (see Sec.~\ref{sec:background}). During the training procedure, we update distribution parameters $\mu_i$ and $\sigma_i$ on the Gaussian distributions for weights and biases as well as parameters $\alpha$ and $\beta$ on the Gamma distribution from which the precision $\tau_n$ is drawn. We collectively refer to all of these model parameters as $\vect{\theta}$.

		The BNN is trained with the NUTS sampler as implemented in PyMC3.\cite{Hoffman2014, Salvatier2016} We start the sampling procedure with 500 samples of burn-in followed by another 1000 samples retaining every tenth sample. This protocol was fixed for all tasks. 

		We can construct an approximation to the kernel density from the distributions of BNN parameters learned from the sampling procedure. In particular, for observations $\mathcal{D}_n$ we compute the kernel densities, which are then used to approximate the objective function. The probability density function of the distribution generated from a single observed parameter point $\vect{x}_k$ can therefore be written in closed form in Eq.~\ref{eq:probabilistic_model_again}, where $\langle \cdot \rangle$ denotes the average over all sampled BNN architectures

		\begin{align}\label{eq:probabilistic_model_again}
    		p_k(\vect{x}) = \left\langle \sqrt{ \frac{\tau_n}{2\pi} } \exp\left[ - \frac{\tau_n}{2} (\vect{x} - \phi_3(\vect{\theta}; \vect{x}_k))^2 \right] \right\rangle_\text{BNN}.
    	\end{align}

    	We formulate the approximation to the objective function as an ensemble average of the observed objective function values $f_k$ taken over the set of computed kernel densities $p_k(\vect{x})$ (see Eq.~\ref{eq:objective_function_approx}). In this ensemble average, each of the constructed distributions $p_k(\vect{x})$ is rescaled by the value of the objective function $f_k$ observed for the parameter point $\vect{x}_k$. The precision $\tau_n$ of the approximation increases with the number of observations and thus converges to the objective function in the limit of infinitely many observations. \\ 

		\begin{align}\label{eq:objective_function_approx}
			\alpha(\vect{x}) = \frac{\sum\limits_{k = 1}^n f_k p_k(\vect{x})}{\sum\limits_{k = 1}^n p_k(\vect{x})}.
		\end{align}

		This approximation to the objective function is faster to evaluate for any given parameter point $\vect{x}$ as the evaluation of this approximation no longer requires a full Bayesian BNN evaluation.

	\subsection{Acquisition function}\label{sec:acquisition_function}


		In the resulting approximation $\alpha(\vect{x})$ we effectively model the probability of a given parameter point $\vect{x}$ to yield the same objective function value $f_k$ of an observed point $\vect{x}_k$. However, the parameter space could contain low density regions, for which the objective function approximation $\alpha(\vect{x})$ is inaccurate.

		Based on these considerations, we propose an acquisition function detailed in Eq.~\ref{eq:proposed_acquisition}. We design the acquisition function from the parameter densities $p_k(\vect{x})$ for observations $\mathcal{D}_n$. The acquisition function differs from the approximation to the objective function (see Eq.~\ref{eq:objective_function_approx}) by an additional term in the numerator and the denominator, which denotes the uniform distribution $p_\text{uniform}(x)$ on the domain. In the numerator, this distribution is scaled by a factor $\lambda$, referred to as the sampling parameter from hereon

		\begin{align}\label{eq:proposed_acquisition}
			\alpha(\vect{x}) = \frac{\sum\limits_{k = 1}^n f_k p_k(\vect{x}) + \lambda p_\text{uniform}(\vect{x})}{\sum\limits_{k = 1}^n p_k(\vect{x}) + p_\text{uniform}(\vect{x})}.
		\end{align}

		The introduced parameter $\lambda$ effectively compares the cumulative height of each rescaled density estimate $p_k(\vect{x})$ to the uniform distribution. While the $p_k(\vect{x})$ are constructed from the knowledge we acquired about the parameter space via objective function evaluations, the uniform distribution is used as a reference to indicate the lack of knowledge in parameter space regions where no information is available yet. The sampling parameter therefore balances between acquired knowledge and the lack of knowledge, which effectively tunes the exploitative and explorative behavior of the algorithm. 

\begin{figure}[!ht]
			\centering
			\includegraphics[width = 0.5\textwidth]{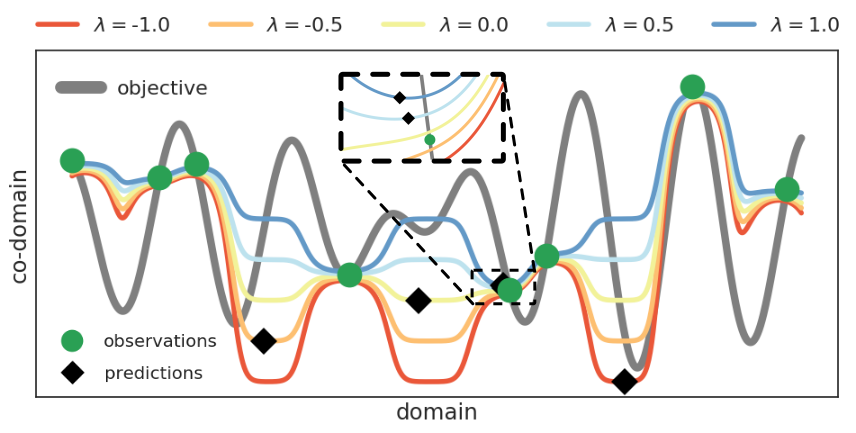}
			\caption{Acquisition functions computed from eight observations (green) of an otherwise unknown objective function (gray). All acquisition functions were constructed with different values of the exploration parameter $\lambda$ as indicated in the legend. New parameter points (black) are proposed based on global minima of the constructed acquisition functions.}
			\label{fig:lambda_influence}
		\end{figure}

		With a large positive value for $\lambda$, \name favors exploitation, while a large negative value favors exploration. When $\lambda = 0$, the acquisition function approximates the objective function itself. Fig.~\ref{fig:lambda_influence} illustrates the behavior of \name on a one-dimensional objective function with different $\lambda$ values. In this example, the acquisition function is constructed from eight observations indicated in green. Acquisition functions which were constructed from a more positive $\lambda$ show low values only in the vicinity of the observation with the lowest objective function value. In contrast, acquisition functions which were constructed from a more negative $\lambda$ show low values far away from any observation. The choice for the value of the exploration parameter $\lambda$ can therefore be directly related to explorative or exploitative behavior.

		From Fig.~\ref{fig:lambda_influence} we see that distinct points in parameter space are proposed based on particular values of the exploration parameter $\lambda$. The best choice of $\lambda$ for a given objective function is \emph{a priori} unknown. However, with the possibility to rapidly construct several acquisition functions with biases towards exploration or exploitation, we can propose multiple parameter points in batches based on different sampling strategies. The newly proposed parameter points are then evaluated on the black-box optimization function in possibly parallel evaluation runs.

	\subsection{Proxy optimization}\label{sec:proxy_optimization}

		New points in parameter space for querying the objective function are proposed based on the global minimum of the acquisition function (see Eq.~\ref{eq:proposed_acquisition}, and Fig. \ref{fig:lambda_influence}). The problem of finding the global minimum of the objective function, which is costly to evaluate, is therefore reduced to searching the global minimum of the function approximation. 

		As a compromise of accuracy and computational cost, we search for the global minimum of the approximating function by uniformly sampling points in the parameter space and then running a gradient based optimizer on half of the proposed samples on the objective function approximation. Unless otherwise noted, all reported results were obtained from proposing $2000$ uniform samples for each dimension of the parameter space and locally optimizing half of the sampled points with the L-BFGS algorithm for at most $20$ optimization steps.\cite{Morales2011}


	\section{Results \& Discussion}\label{sec:results}

	In this section we report the performance of \name and compare it to two frequently used global optimization packages built on Bayesian optimization. The ``spearmint'' software package performs Bayesian optimization using GPs and the predictive entropy acquisition function.\cite{Snoek2012, Snoek2014} The SMAC software employs RF models.\cite{Hutter2011, Hutter2012, smac2017} 

	The performance of each of these three optimization algorithms is evaluated on a set of 15 qualitatively different benchmark functions, which are continuous and convex, non-convex or discrete with possibly multiple local minima. Nine of the employed functions, which have a continuous co-domain, are well-established benchmarks for global optimization benchmarks. Additionally, we design six difficult cases with discrete co-domain. A complete list of the employed objective functions as well as their global minima is provided in the supplementary information (see Sec.~\ref{sec:supp_loss_functions}). 

	We performed $20$ independent runs initialized with different random seeds, unless noted otherwise. During each optimization run we record the lowest achieved objective function value after each iteration. We compare the averaged lowest achieved objective function values by relating to results from simple random searches. Each random search was run for $10^4$ objective function evaluations and results were averaged over $50$ independent runs initialized with different random seeds. The average lowest achieved objective function values of the random search runs are summarized in the supplementary information (see Tab.~\ref{tab:random_search_results}).

	\subsection{Two dimensional benchmarks}\label{sec:simple_benchmarks}

		We start our discussion with a benchmark on the test functions mentioned above on two dimensional parameter spaces. \name was set up with three different values for the sampling parameter, $\lambda \in \{-1, 0, 1\}$, to assess the effectiveness of a particular parameter choice. Lowest objective function values achieved by each of the three considered optimization algorithms were recorded for $20$ independent runs initialized with a common set of random seeds. 

\begin{figure}[!ht]
			\centering
			\includegraphics[width = 1.0\columnwidth]{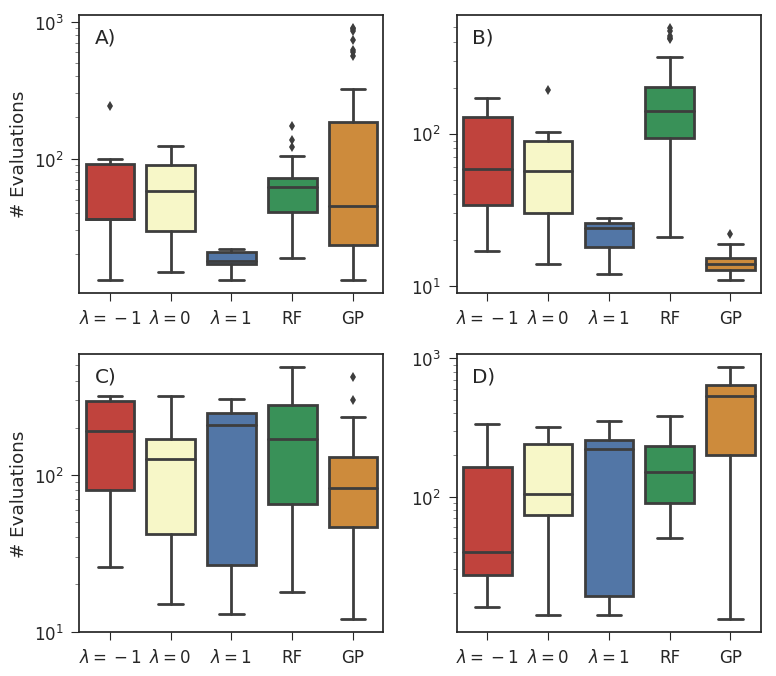}
			\caption{Number of objective function evaluations required to reach objective function values lower than the average lowest achieved values of random searches with $10^4$ evaluations for \name ($\lambda \in \{-1, 0, 1\}$), RFs and GPs. Results are reported for the Ackley (A), Dejong (B), Schwefel (C) and dAckley (D) objective functions. Details on the benchmark functions are provided in the supplementary information Sec.~\ref{sec:supp_loss_functions}}
			\label{fig:simple_benchmarks}
		\end{figure}

		In Fig.~\ref{fig:simple_benchmarks} we report the number of objective function evaluations required by each of the optimization algorithms to reach an objective function value lower than the average lowest value found in random searches within $10^4$ evaluations (see Tab.~\ref{tab:random_search_results}). Optimization traces for these runs on all 15 objective functions are reported in the supplementary information (see Sec.~\ref{sec:benchmark_results}).

		We find that GP optimization, as implemented in spearmint, generally quickly finds the global minimum if the objective function is strictly convex. In contrast, RF optimization, as implemented in SMAC, quickly finds the global minimum of objective functions with a discrete co-domain.

		The performance of \name varies with different values of the sampling parameter $\lambda$. When favoring exploitation over exploration, i.e. $\lambda > 0$ the algorithm performs better if the objective function features narrow and well defined funnels (e.g. Ackley in Fig.~\ref{fig:simple_benchmarks}a or Schwefel in Fig.~\ref{fig:simple_benchmarks}c). With this choice for the sampling parameter, the algorithm is slightly biased towards exploring the local region around the current optimum. This behavior, however, is unfavorable in other cases, for instance when the objective function has a discrete co-domain (e.g. dAckley in Fig.~\ref{fig:simple_benchmarks}d). Since parameter points in the vicinity to the current optimum likely yield the same value if the objective function is discrete or quasi-discrete, \name performs better on such objective functions when favoring exploration over exploitation, i.e. $\lambda < 0$.

	\subsection{Batch optimization}\label{sec:batch_optimization}

		The dependence of the performance of \name on the sampling parameter $\lambda$ could be eliminated by marginalizing over this parameter. Marginalization over the sampling parameter would effectively average out the advantageous effects of a bias towards exploitation for some objective functions and towards exploration for other objective functions. 

		The shape of the objective function is \emph{a priori} unknown, so suitable choices of the sampling parameter cannot be determined beforehand, hence marginalization appears to be a solution. However, since the sampling parameter can be directly related to the explorative and exploitative behavior of the algorithm (see Sec.~\ref{sec:acquisition_function}) we continue with investigating a different approach to take full advantage of the sampling policy. We suggest to propose new parameter points based on a number of different sampling parameter values to keep the advantages of favored exploration and favored exploitation. 

		Given a set of observations $\mathcal{D}_n$ the construction of several approximations to the objective function with different values of $\lambda$ is computationally cheap. We can therefore easily suggest multiple new parameter points in batches at each optimization iteration, which are proposed from more explorative and more exploitative parameter values, at almost no additional cost. With the observations on the simple benchmarks in Sec.~\ref{sec:simple_benchmarks} we would expect a synergistic effect of this batch optimization over sequential optimization with a single sampling parameter value. As parameter points can be proposed with both a bias towards exploration and a bias towards exploitation, we expect the number of required objective function evaluations for reaching the global optimum to decrease. In addition, suggesting a batch of parameter points in one optimization step allows for the parallel evaluation of all proposed points, which accelerates the optimization process. 

		We demonstrate the aforementioned synergistic effect by running the optimization algorithm on the previously studied objective functions (see Fig.~\ref{fig:simple_benchmarks}) with a different number of parameter points proposed per batch. Both, spearmint and SMAC allow for a parallel evaluation of a given objective function. With the spearmint package, multiple samples are generated by marginalizing over the possible outcomes of currently running experiments. The SMAC package allows for parallel objective function evaluations by running multiple regressor instances, which share a common set of objective function evaluations.

		Tab.~\ref{tab:swarm_exploration_results} summarizes the results for a simple batch optimization experiment on a selected set of objective functions. We again report the number of evaluations needed by each optimizer to reach an objective function value lower than the lowest average value found in random search with $10^4$ iterations. \name was run with sampling parameter values evenly spread out over the $[-1, 1]$ interval. Note, that the evaluation of a batch with $p$ samples was counted as $p$ evaluations of the objective functions. Traces of the lowest achieved objective function values are also depicted in the supplementary information (see Sec.~\ref{sec:swarm_exploration}).

		\begin{table}[!ht]
			\setlength{\tabcolsep}{4pt}
			\centering
			\begin{tabular}{lcrrrrrr}
				\toprule
				Method & $p$ & Ackley & Dejong & Schwefel & dAckley \\
				\midrule
				\multirow{4}{*}{PHOENICS}       & $1$ & \error{39}{5} & \error{55}{\phantom{1}4} & \error{66}{\phantom{1}4} & \error{85}{\phantom{1}8} \\
									       & $2$ & \smerr{26}{3} & \error{44}{\phantom{1}2} & \error{58}{\phantom{1}6} & \error{72}{\phantom{1}6} \\
								           & $4$ & \error{43}{1} & \smerr{36}{\phantom{1}1} & \error{48}{\phantom{1}8} & \smerr{22}{\phantom{1}2} \\
								           & $8$ & \error{69}{2} & \error{61}{\phantom{1}2} & \smerr{47}{\phantom{1}2} & \error{40}{\phantom{1}3} \\
				\midrule
				\multirow{4}{*}{RF}      & $1$ & \smerr{185}{2} & \smerr{250}{\phantom{1}9} & \smerr{245}{\phantom{1}3} & \smerr{143}{\phantom{1}6} \\
									       & $2$ & \error{191}{5} & \error{349}{11}           & \error{277}{\phantom{1}6} & \error{166}{\phantom{1}5} \\
									       & $4$ & \error{212}{6} & \error{713}{32}           & \error{394}{10} & \error{178}{\phantom{1}5} \\
									       & $8$ & \error{246}{8} & \error{906}{47}           & \error{446}{17} & \error{225}{28} \\
				\midrule
				\multirow{4}{*}{GP} & $1$ & \smerr{43}{3} & \error{13}{\phantom{1}1} & \smerr{93}{\phantom{1}2} & \smerr{100}{12} \\
										   & $2$ & \error{45}{3} & \smerr{12}{\phantom{1}1} & \error{94}{\phantom{1}3} & \error{107}{19} \\
										   & $4$ & \error{46}{2} & \error{12}{\phantom{1}1} & \error{96}{\phantom{1}2} & \error{126}{\phantom{1}6} \\
										   & $8$ & \error{57}{3} & \error{16}{\phantom{1}1} & \error{95}{\phantom{1}2} & \error{163}{20} \\
				\bottomrule
			\end{tabular}
			\caption{Average lowest achieved errors of the three global optimization algorithms compared in this study. Averages were taken over $20$ runs with different random seeds. For each optimizer we report the lowest achieved errors for runs in which a different number of points $p$ were proposed in each training iteration. Lowest numbers of evaluations required by each optimization algorithm are indicated in bold. }
			\label{tab:swarm_exploration_results}
		\end{table}

\begin{figure}[!ht]
			\centering
			\includegraphics[width = 1.0\columnwidth]{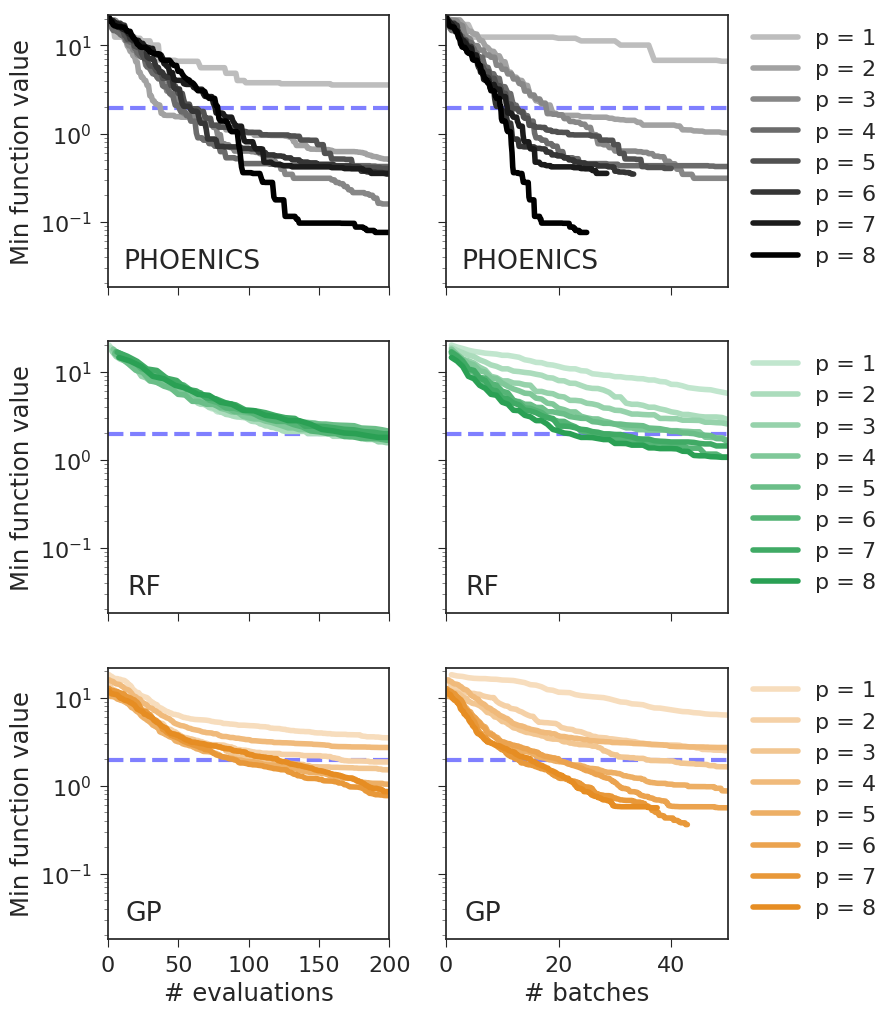}
			\caption{Average minimum objective function values for the Ackley function achieved in $20$ independent runs of the three optimization algorithms studied in this work: our optimizer (PHOENICS), spearmint (GP) and SMAC (RF). For each run a different number of proposed samples $p$ was evaluated in parallel. Minimum achieved objective function values are reported with respect to the total number of objective function evaluations and the number of evaluated batches. The dashed blue lines denote the minimum achieved error after $10^4$ of random search for reference. }
			\label{fig:swarm_behavior_one_loss}
		\end{figure}

		The behavior of the three studied optimization algorithms under parallel optimization on the Ackley objective function is illustrated in Fig.~\ref{fig:swarm_behavior_one_loss}. In this figure we depict the minimum achieved objective function values for different runs with a different number of parallel evaluations of the objective function averaged over $20$ independent runs. The minimum achieved objective function values are presented per number of objective function evaluations (left panel) and per batch evaluation (right panel).

		We find that both spearmint and SMAC achieve low objective function values in fewer batches with an increasing number of points $p$ proposed in each batch. While increasing the number of samples proposed per batch initially significantly improves the performance with respect to the number of proposed batches, this advantageous effect quickly levels off until there is no significant improvement beyond six samples per batch. However, when comparing the minimum achieved objective function values with respect to the total number of objective function evaluations, we did not observe any significant difference between runs with a different number of samples proposed per batch.

		In contrast, \name shows a different behavior. Our algorithm does not only reach lower objective function values in a fewer number of batches when proposing more samples per batch, but also shows a better performance when considering the total number of function evaluations. This synergistic effect demonstrates that \name indeed benefits from proposing points in batches even in cases in which proposed samples are evaluated sequentially.

		The performance improvement of \name when proposing parameter points in batches at each optimization iteration is demonstrated on all 15 considered objective functions in the supplementary information (see Sec.~\ref{sec:swarm_exploration}). We ran our optimizer with four points per batch, which are proposed from sampling parameter values evenly spaced across the $[-1, 1]$ interval. All four proposed parameter points are then evaluated before we started another optimization iteration. For this particular batching protocol, we find that \name outperforms RF based optimization on all benchmark functions and GP based optimization on 12 out of 15 benchmark functions. If and only if the objective function is convex, GP optimization finds lower objective function values. In addition, we observe the aforementioned synergistic effect of batch optimization for 12 out of 15 benchmark functions. Despite reducing the number of optimization iterations by a factor of 4, the achieved objective function values were found to be lower than values achieved in sequential optimizations with all three considered fixed sample parameter value.

\begin{figure}[!ht]
			\includegraphics[width = 1.0\columnwidth]{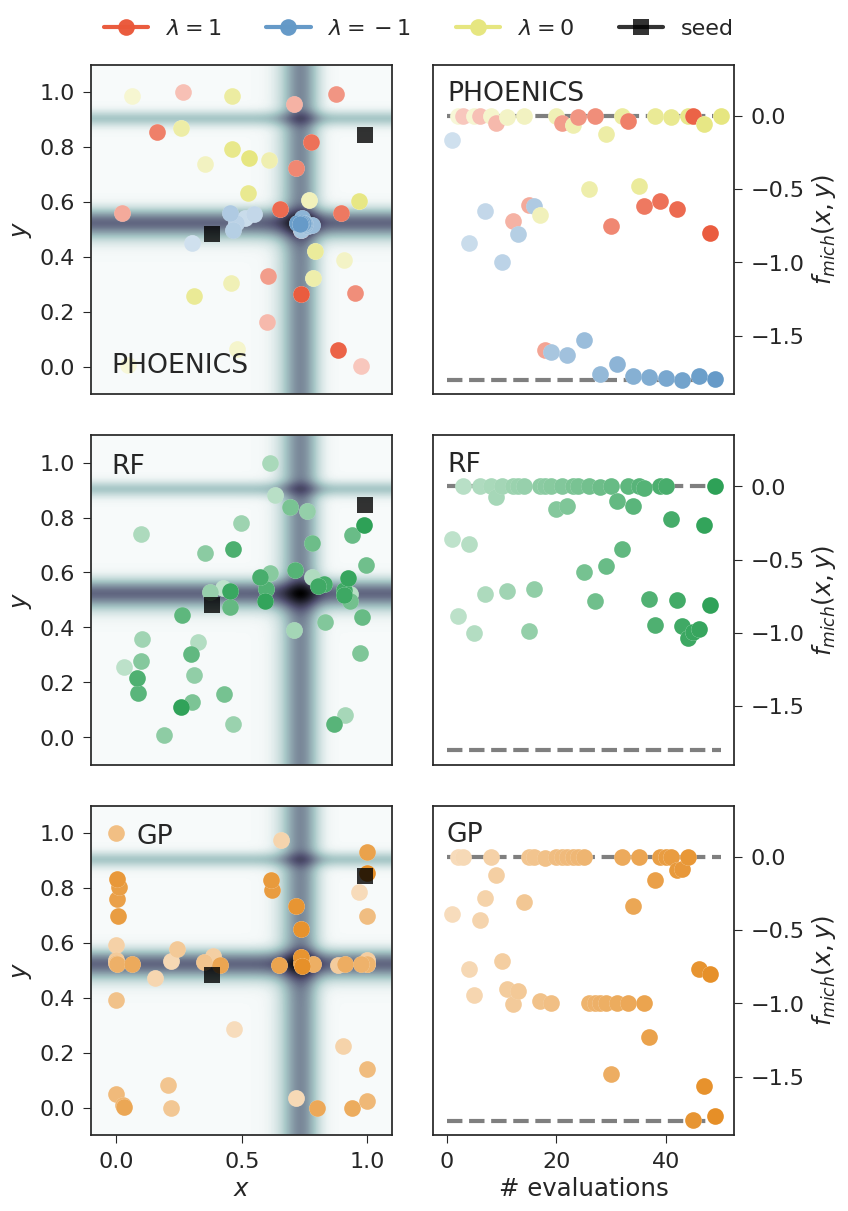}
			\caption{Progress of sample optimization runs of the three studied optimization algorithms on the two dimensional Michalewicz function. \name proposed a total of three samples per batch, which were then evaluated in parallel. Each sample was suggested based on a particular value of the exploration parameter $\lambda \in \{-1, 0, 1\}$. Left panels illustrate the parameter points proposed at each optimization iteration while right panels depict the achieved objective function values. Depicted points are more transparent at the beginning of the optimization and more opaque towards the end. Starting points for the optimization runs are drawn as black squares. }
			\label{fig:michalewicz_samples}
		\end{figure}

		We suggest that this improved performance of the algorithm is due to the trade-off between exploration and exploitation: while some of the samples proposed in one batch were suggested from parameter values $\lambda$ which favor exploration, others are proposed for exploitation and locally exploring the region close to the currently lowest discovered objective function value. The exploration samples therefore systematically sample the parameter space and ensure that the algorithm does not get stuck in local minima, while the exploitation samples explore the local environment of the current global minimum. This sampling behavior is illustrated in Fig.~\ref{fig:michalewicz_samples} for the Michalewicz function. 

		The optimization runs on the Michalewicz function depicted in Fig.~\ref{fig:michalewicz_samples} were all started from the same two random samples illustrated in black for all three investigated optimization algorithms. Bayesian optimization based on GPs as implemented in spearmint (lower panels) tends to sample many parameter points close to the boundaries of the domain space in this particular example. RF optimization as implemented in SMAC (central panels), however, shows a higher tendency of exploring the parameter space. 

		\name (upper panels) starts exploring the space and quickly finds a local minimum in vicinity of one of the initial samples. After finding this local minimum, samples which are proposed based on a more exploitative (positive) value of the sampling parameter $\lambda$ explore the local environment of this local minimum while samples proposed from more explorative (negative) values of $\lambda$ explore the entire parameter space. As soon as the exploration points find a point in parameter space with a lower value of the objective function, the exploitation points jump to this new region in parameter space and locally explore the region around the current best to quickly converge to the global minimum. 

		Overall we have demonstrated that the value of the sampling parameter $\lambda$ in the proposed acquisition function clearly influences the behavior of the optimization procedure towards a more explorative behavior for more negative values of this parameter and a more exploitative behavior for more positive parameter values. Batched optimization improves the performance of \name even in terms of total objective function evaluations and reduces the number of required optimization iterations.

	\subsection{Increasing the number of dimensions}

		Real-life chemical problems are frequently concerned with more than just two parameters. Chemical reactions can be influenced by environmental conditions and experimental device settings, and computational studies frequently employ parameters to describe the system of interest. As such, this section focuses on the performance of \name in parameter spaces with dimensions $k > 2$. 

		When moving towards higher dimensional spaces, the chance for a uniformly sampled point to be close to the boundaries of the considered $k$-dimensional hypercube increases, i.e. most sampled points will be close to the edges. Likewise, the volume of the region close to the global optimum decreases with respect to the total volume of the hypercube. Thus, we expect \name, RF and GP to perform worse for higher dimensional spaces. 

		We evaluated the performance of the three considered optimization algorithms on the subset of objective functions already considered in Sec.~\ref{sec:simple_benchmarks}, but now successively increase the dimensionality of the parameter space from two to $20$. Based on the results on batch optimization (see Sec.~\ref{sec:batch_optimization}) we ran GP optimization and RF optimization with one point per batch and the optimization algorithm introduced in this study with four points per batch on each considered benchmark function. Exploration parameter values were chosen to be evenly spaced across the $[-1, 1]$ interval. 

\begin{figure}[!ht]
			\centering
			\includegraphics[width = 0.5\textwidth]{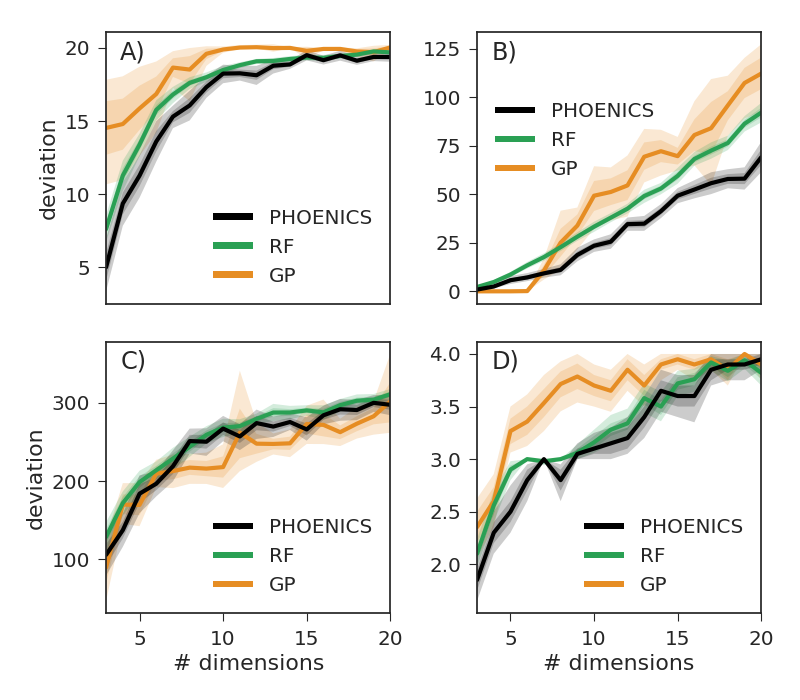}
			\caption{Average deviations taken over $20$ independent runs between the lowest encountered objective function value and the global minimum achieved after $200$ objective function evaluations for different parameter set dimensions. Results are reported for Ackley (A), Dejong (B), Schwefel (C) and dAckley (D). Uncertainty bands illustrate bootstrapped estimates of the deviation of the means with one and two standard deviations.}
			\label{fig:dimensionality_scan}
		\end{figure}

		For better comparisons we report the average deviation of the lowest encountered objective function value from the global minimum of each function taken over $20$ independent optimization runs. Average deviations achieved by each of the optimization algorithms after $200$ objective function evaluations are depicted in Fig.~\ref{fig:dimensionality_scan}.

		We observe that \name maintains its rapid optimization properties for a variety of different objective functions even when increasing the number of dimensions. In the case of the Ackley function (Fig.~\ref{fig:dimensionality_scan}a) \name appears to find and explore the major funnel close to the global optimum faster than the other two optimization algorithms regardless of the number of dimensions. The paraboloid (Fig.~\ref{fig:dimensionality_scan}b) is an easy case for the GP in low dimensions, but is optimized the fastest by \name when considering parameter spaces with seven or more dimensions. No major differences are observed for the Schwefel function (Fig.~\ref{fig:dimensionality_scan}c). However, in the case of a discrete objective function (Fig.~\ref{fig:dimensionality_scan}d) \name seems to have a slight advantage over the other two optimizers for lower dimensions and performs about as well as random forest optimization for higher dimensions.


	\section{Applications to chemistry}\label{sec:applications}

	\name was shown to rapidly find the global optima of analytic objective functions used for benchmarking global optimization algorithms (see Sec.~\ref{sec:results}). In this section, we demonstrate its performance on the Oregonator, a model system of a chemical reaction described by a set of non-linear coupled differential equations.\cite{Field1974}

	Most chemical reactions lead to a steady-state, i.e. a state in which the concentrations of involved compounds are constant in time. While chemical reactions described by a single differential equation always feature such a steady-state, more complicated dynamics phenomena can arise for reactions described by sets of nonlinear coupled differential equations. With the right choice of parameters, such differential equations may have a stable limit cycle, leading to periodic oscillations in the concentrations of involved compounds.\cite{Kampen1992, Zhabotinsky1991} With other parameter choices, however, the same chemical system might feature an attractive fixed point instead.

	One of the earliest discovered reactions featuring a stable limit cycle for a set of reaction conditions is the Belousov-Zhabotinsky reaction.\cite{Degn1967, Zhabotinsky1967} This network of chemical reactions involves temporal oscillations of [Ce$^\text{IV}$] and [Ce$^\text{III}$]. The entire reaction network can be written as a set of three subreactions listed in Reaction~\ref{reaction:belousov-zhabotinsky}. For details on the mechanism we refer to a brief summary in the supplementary information (see Sec.~\ref{sec:belousov_zhabotinski}) as well as to the literature.\cite{Zhabotinsky1967, Field1972, Field1974, Gyorgyi1990, Field1973}

	Models at different levels of complexity have been developed to describe the dynamic behavior of the Belousov-Zhabotinsky reaction.\cite{Field1973, Bar1985, Kampen1992, Voorsluijs2017} One of the simplest models of this reaction is the Oregonator.\cite{Field1974} The Oregonator consists of a set of three coupled first order non-linear differential equations for three model compounds $X$, $Y$ and $Z$, which are shown in Eqs.~\ref{eq:oregonator_0} to \ref{eq:oregonator_2}. These equations involve five reaction constants $k_i$, a stoichiometric factor $f$, determined by the prevalence of one subreaction over another subreaction, and the concentration of two additional chemical compounds $A$ and $B$. A map of Eqs.~\ref{eq:oregonator_0} to \ref{eq:oregonator_2} to Reaction~\ref{reaction:belousov-zhabotinsky} is outlined in Ref.~\cite{Field1974}

	\begin{align}
		\frac{dX}{dt} &= k_1 A Y - k_2 X Y + k_3 B X - 2 k_4 X^2,  \label{eq:oregonator_0} \\
		\frac{dY}{dt} &= -k_1 A Y - k_2 X Y + f k_5 Z, \label{eq:oregonator_1} \\
		\frac{dZ}{dt} &= k_3 B X - k_5 Z. \label{eq:oregonator_2} 
	\end{align}

	The set of differential equations in the Oregonator can be reduced into a dimensionless form, such that the number of correlated parameters is reduced to a smaller set of independent parameters. This reduced version of the Oregonator, shown in Eqs.~\ref{eq:red_oregonator_0} to \ref{eq:red_oregonator_2}, includes three dimensionless variables $\alpha$, $\eta$ and $\rho$, which describe the concentration of chemical species, and four dimensionless reaction constants $q$, $s$, $w$ and $f$.

	\begin{Reaction*}[!ht]
		\begin{align}
			\ce{
				2Br- + BrO3- + 3H+ + 3CH2(COOH)2 &-> 3BrCH(COOH)2 + 3H2O \tag{A}\label{reaction:A}\\
				BrO3- + 5H+ + 4Ce^{III} &-> 4Ce^{IV} + HOBr + 2 H2O \tag{B}\label{reaction:B}\\
				2Ce^{IV} + 2CH2(COOH)2 + BrCH(COOH)2 + HBrO + 2H2O &-> 2Ce^{III} + 2Br- + 3HOCH(COOH)2 + 4H+ \tag{C}\label{reaction:C}
			}
		\end{align}
		\caption{Subreactions of the Belousov-Zhabotinsky reaction.\cite{Kuhnert1987}}
		\label{reaction:belousov-zhabotinsky}
	\end{Reaction*}

	\begin{align}
		\frac{d\alpha}{d\tau} &= s (\eta - \eta \alpha + \alpha - q \alpha^2), \label{eq:red_oregonator_0} \\
		\frac{d\eta}{d\tau} &= s^{-1} ( - \eta - \eta \alpha + f \rho), \label{eq:red_oregonator_1} \\
		\frac{d\rho}{d\tau} &= w (\alpha - \rho). \label{eq:red_oregonator_2}
	\end{align}

	\name is used to reverse engineer the set of reaction parameters consisting of three initial conditions and four reaction constants from the concentration traces computed in the original publication.\cite{Field1974} The goal is twofold: (i) find a set of parameters for which the dynamical behavior qualitatively agrees with the behavior of the target, i.e. find chemical oscillations and (ii) fine tune this set of parameters such that we reproduce the dynamical behavior on a quantitative level to accurately predict the concentration traces for the involved compounds.

	We implemented a fourth-order Runge Kutta integrator with adaptive time stepping to compute the concentration traces for a given set of reaction parameters. For each proposed set we use this integrator to evolve the differential equations and obtain concentrations on a regular grid via cubic spline interpolation. The integrator was run for a total of $10^7$ integration steps covering $12$ full oscillation periods for the target parameter set. Numerical values for the target parameter set as well as the boundaries of the search space are provided in the supplementary information (see Sec.~\ref{sec:belousov_zhabotinski}). The interpolated concentration traces are compared to the target traces by calculating an objective loss function based on the euclidian distance between the points in time at which a concentration trace reaches a dimensionless concentration value of $100$. 

	\begin{figure}[!ht]
		\centering
		\includegraphics[width = 1.0\columnwidth]{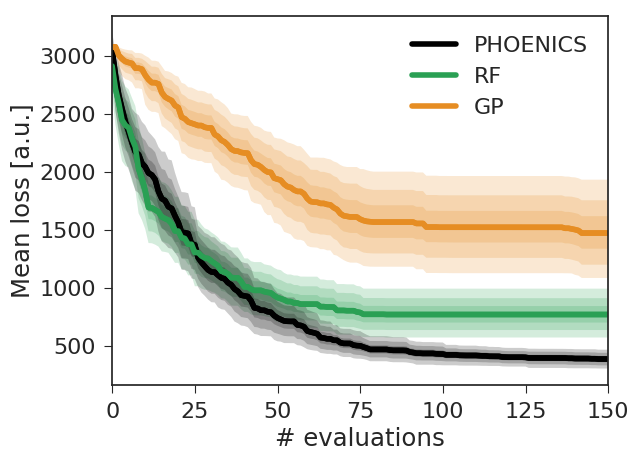}
		\caption{Average achieved losses for finding reaction parameters of the reduced Oregonator model achieved by the three optimization algorithms employed in this study. Correct periodicities of the concentration traces are achieved for losses lower than $500$. Uncertainty bands illustrate bootstrapped deviations on the mean for one and two standard deviations. }
		\label{fig:oregonator_losses}
	\end{figure}

	\name was run in parallel proposing four samples per batch with $\lambda$ equally spaced on the $[-1, 1]$ interval. We compare the performance to GP optimization in spearmint and RF optimization in SMAC (see Sec.~\ref{sec:results}). Each of the three optimization algorithms was used in $50$ independent optimization runs for $150$ iterations. Average achieved losses for all three optimization algorithms are displayed in Fig.~\ref{fig:oregonator_losses}. Loss function values between $300$ and $500$ indicate that the periodicity of the predicted concentration traces resembles the periodicity of the target traces, i.e. the predicted traces qualitatively agree with the target. Quantitative agreement, i.e. matching traces, is only achieved for loss values lower than about $100$. Examples for concentration traces yielding different losses are presented in the supplementary information (see Fig.~\ref{fig:oregonator_trace_plot_collection}).

\begin{figure}[!ht]
		\centering
		\includegraphics[width = 1.0\columnwidth]{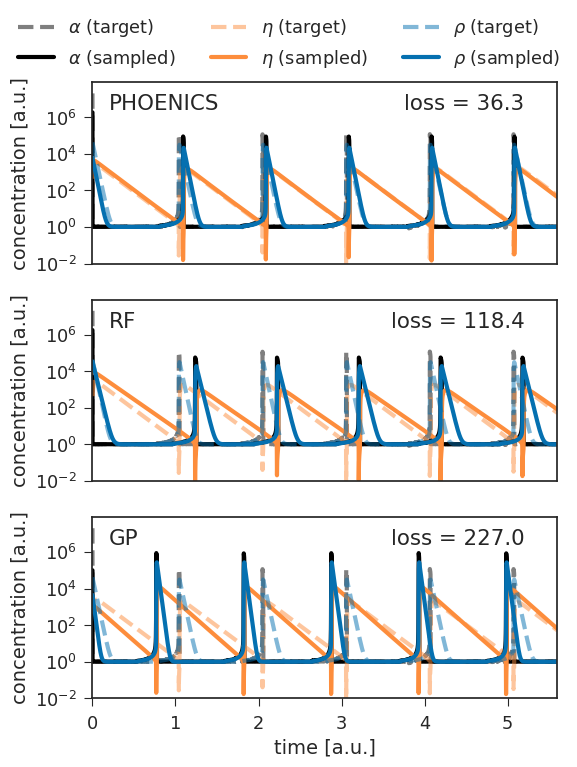}
		\caption{Time traces of dimensionless concentrations of compounds in the \emph{Oregonator} model. Target traces are depicted with solid, transparent lines while predicted traces are shown in dashed, opaque lines. Traces were simulated for a total of $12$ dimensionless time units, but are only shown for the first six time units for clarity. }
		\label{fig:oregonator_traces}
	\end{figure}

	Fig.~\ref{fig:oregonator_traces} shows concentration traces associated to the lowest loss achieved by each of the three optimization algorithms across all $50$ independent runs. \name is the only algorithm reproducing qualitatively and quantitatively target dynamic behavior within $150$ optimization iterations. RF optimization only finds parameter sets which qualitatively agree with the target. GP optimization finds only in rare occasions concentration traces in qualitative agreement with the target.

	\section{Conclusion}

	In this work we introduced \namecomma a novel algorithm for the global optimization of expensive to evaluate black-box objective functions. Our probabilistic optimizer combines Bayesian optimization with conceptual aspects of Bayesian Kernel Density estimation. Through an exhaustive benchmark study, we showed that \name improves over existing global optimization methods.  

	We formulate an inexpensive acquisition function balancing the explorative and exploitative behavior of the algorithm. This acquisition function enables intuitive sampling policies for an efficient parallel search of global minima. By leveraging synergistic effects from running multiple sampling policies in batches, the performance of the algorithm improves, and requires a reduced total number of objective function evaluations.

	Benchmark results were compared to popular Bayesian optimization methods based on GPs and RFs. Unlike GP and RF, \name appears to be less sensitive to the nature of the co-domain. Notably, \name was shown to outperform RF in all 15 test functions, and to outperform GP in all 12 non-convex test functions. When moving towards higher dimensional spaces, we observed that \name maintains its rapid optimization properties for a variety of different objective functions.

	We illustrated the capabilities of our algorithm for an inverse design problem on the Oregonator, a practical example for a model system describing a complex chemical reaction network. We demonstrated that only \name was able to reproduce the qualitative and quantitative target dynamic behavior of the nonlinear reaction dynamics.  

	We believe that \name has the potential to be applied to a wide range of applications, from optimization of reaction conditions and material properties, over control of robotics systems, to circuit design for quantum computing.\cite{Peruzzo2014, Romero2017}

	All in all, we recommend \name for an efficient optimization of scalar, possibly non-convex, black-box unknown objective functions. 


	\section*{Acknowledgments}

	We thank Dr.~S.~K.~Saikin for fruitful discussions and helpful comments. F.H. was supported by the Herchel Smith Graduate Fellowship. L.M.R and A.A.G were supported by the Tata Sons Limited - Alliance Agreement (A32391). C.K. and A.A.G were supported by the National Science Foundation under award number CHE-1464862. All computations reported in this paper were completed on the Odyssey cluster supported by the FAS Division of Science, Research Computing Group at Harvard University.

		\phantomsection\addcontentsline{toc}{section}{\refname}
		\putbib[main]	
	\end{bibunit}

	\clearpage
	\newpage

	\begin{bibunit}[unsrt]
		\onecolumngrid
		\setcounter{subsection}{0}

\section*{Supplementary information}

	\subsection{Analytic objective functions}\label{sec:supp_loss_functions}

		We benchmarked the performance of the global optimization algorithm \name on a total of 15 analytic objective functions. Nine of these objective functions have a continuous co-domain and are commonly used to benchmark algorithms developed for unconstrained optimization problems. The remaining six benchmark functions have a discrete co-domain and were specifically designed for this study. We provide Python implementations for all 15 benchmark functions on a GitHub repository.\cite{githubRepo}

\begin{figure}[!ht]
			\centering
			\includegraphics[width = 1.0\textwidth]{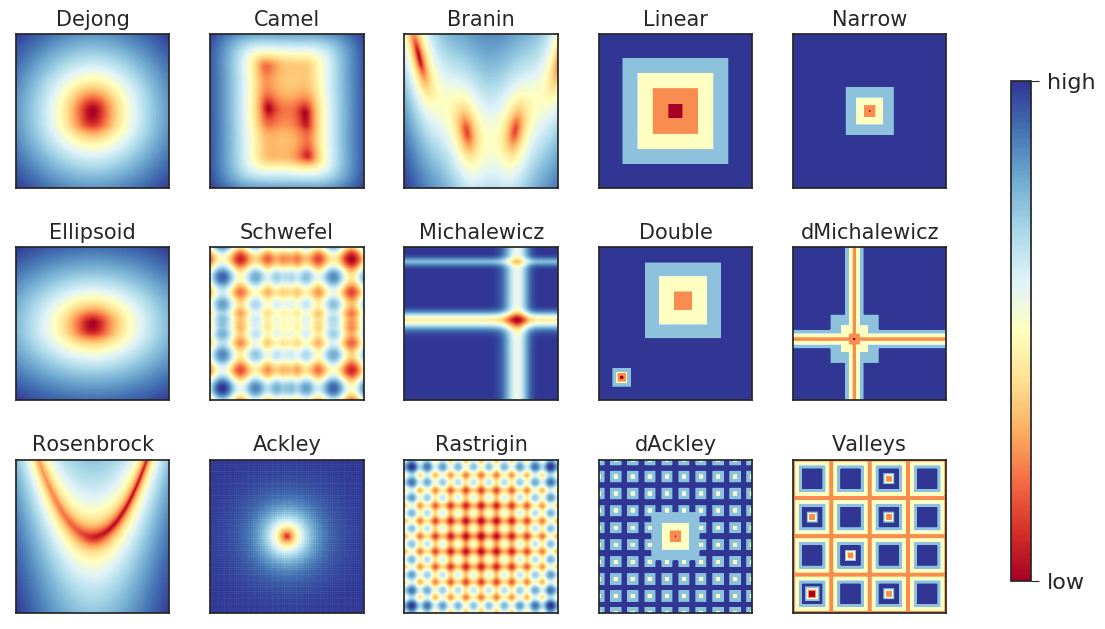}
			\caption{Contour plot of the two dimensional instances of the objective functions with discrete co-domain used for benchmarking \namecomma RF and GP optimizaters. Python implementations of these objective functions are available on GitHub.\cite{githubRepo}}
			\label{fig:loss_functions}
		\end{figure}

		The nine analytic benchmark functions with continuous co-domain were chosen based on their different features: they differ in the number of local minima, number of global minima and the dimensionality of the parameter space for which they are defined. Details are summarized in Tab.~\ref{tab:loss_function_features}. Fig.~\ref{fig:loss_functions} displays contour plots of all objective functions for a two dimensional parameter space.

		All of the benchmark functions with discrete co-domain project the parameter space onto integer values in the $[0, 4]$ interval. Contour plots for these benchmark functions are displayed in Fig.~\ref{fig:loss_functions} for a two dimensional parameter space, although all of the six functions generalize to higher dimensions as well. Note, that the global minimum of these functions is not a unique point in parameter space, but rather an entire region.

		\begin{table}[!ht]
			\centering
			\begin{tabular}{llrc}
				\toprule 
				Objective function   & Domain range      & Global minimum (2d) & generalizable \\
				\midrule  
				Ackley               & $x_i \in [-32, 32]$   & $0.00$       & yes \\
				Branin               & $x_i \in [-5, 15]$    & $0.397887$   & no  \\
				Camel                & $x_i \in [-3, 3]$     & $-1.0316$    & no  \\
				Dejong               & $x_i \in [-5, 5]$     & $0.00$       & yes \\  
				Ellipsoid            & $x_i \in [-5, 5]$     & $0.00$       & yes \\
				Michalewicz          & $x_i \in [0, 3]$      & $-1.801$     & no  \\
				Rastrigin            & $x_i \in [-5, 5]$     & $0.00$       & yes \\
				Rosenbrock           & $x_i \in [-2, 2]$     & $0.00$       & yes \\
				Schwefel 	         & $x_i \in [-500, 500]$ & $-418.9829d$ & yes \\
				\bottomrule
			\end{tabular}
			\caption{Analytic objective functions with continuous co-domain used for benchmarking \name, RF and GP optimizaters. Python implementations of these objective functions are available on GitHub.\cite{githubRepo}}
			\label{tab:loss_function_features}
		\end{table}

	\subsection{Random search results}\label{sec:random_search_results}

		The employed analytic benchmark functions (see Sec.~\ref{sec:supp_loss_functions}) differ drastically in their shape. While some functions, like the Ackley function or the Schwefel function feature narrow local funnels around their global minima, other functions such as Branin and Dejong have rather broad funnels. The benchmark functions also differ greatly in the range of their co-domain spaces. Instead of reporting the deviation of the current best to the global minimum during an optimization run we therefore compare the optimization algorithms to random searches on the same objective functions. 

		In one random search run we evaluated each objective function $10^4$ times at positions uniformly sampled from the domain space. The lowest achieved function value averaged over $100$ random searches with different random seeds serves as the benchmark value for each optimization algorithm. For each algorithm, we record the number of function evaluations needed to discover a point in parameter space which yields a function value lower than the average lowest value encountered in the random searches. The average lowest values for all objective functions in two dimensions are summarized in Tab.~\ref{tab:random_search_results}. 

		\begin{table}[!ht]
			\centering
			\begin{minipage}[t]{0.45\textwidth}
				\vspace{0pt}
				\begin{tabular}{lrr}
					\toprule
					Loss function & Global minimum & Lowest encounter \\
					\midrule
					Ackley          &     $0.00$ & $1.942$ \\
					Branin          & $0.397887$ & $0.406$ \\
					Camel           &  $-1.0316$ & $-1.028$ \\
					Dejong          &     $0.00$ & $2.560\cdot 10^{-3}$ \\
					Ellipsoid       &     $0.00$ & $3.467\cdot 10^{-3}$ \\
					Michalewicz     &   $-1.801$ & $-1.794$ \\
					Rastrigin       &     $0.00$ & $4.498\cdot 10^{-1}$ \\
					Rosenbrock      &     $0.00$ & $4.718\cdot 10^{-3}$ \\
					Schwefel        & $-837.978$ & $-834.688$ \\
					\bottomrule
				\end{tabular}
			\end{minipage}
			\begin{minipage}[t]{0.45\textwidth}
				\vspace{0pt}
					\begin{tabular}{lrr}
					\toprule
					Loss function & Global minimum & Lowest encounter \\
					\midrule
					Linear funnel     & $0$ & $0.00$ \\
					Narrow funnel     & $0$ & $0.66$ \\
					Double well       & $0$ & $0.36$ \\
					Disc. Ackley      & $0$ & $0.66$ \\
					Disc. Michalewicz & $0$ & $0.64$ \\
					Disc. Valleys     & $0$ & $0.18$ \\
					\bottomrule
				\end{tabular}
			\end{minipage}
			\caption{Lowest achieved function values averaged over $100$ independent random searches with $10^4$ function evaluations per run. The reported values were used to benchmark \namecomma RF and GP optimizers.}
			\label{tab:random_search_results}
		\end{table}

	\subsection{Precision}\label{sec:supp_shrinkage}

		In this work we propose to approximate a given objective function with a prior constructed from Gaussian distributions. Given a set of $n$ observations $\mathcal{D}_n$ of pairs of parameter points and corresponding objective function values, the approximation to the objective function is constructed from $n$ Gaussians. The locations of the Gaussian distributions are drawn from a BNN trained to predict the locations of observed parameter points. Precisions $\tau$ of these Gaussians, however, are drawn from a Gamma distribution parametrized via a probability density function as presented in Eq.~\ref{eq:gamma_distribution}, where $\Gamma$ denotes the Gamma function. 

		\begin{align}\label{eq:gamma_distribution}
			f(x; \alpha, \beta) = \frac{\beta^\alpha x^{\alpha - 1} e^{-\beta x}}{\Gamma(\alpha)} \qquad \text{for } x > 0 \text{ and } \alpha,\beta > 0
		\end{align}

		The prior for this Gamma distribution is chosen to be $\alpha = 12 n^2$ and $\beta = 1$ for a given set $\mathcal{D}_n$ of $n$ observations. With this choice of hyperparameters the expectation value for the prediction is $\langle \tau \rangle = 12 n^2$. The prefactor $12$ in the expectation value of the precision ensures that in the case of only a single observation the standard deviation of the approximating Gaussian distribution matches the standard deviation of a uniform distribution along a single dimension of the unit hypercube. Starting out from an uninformative uniform prior we therefore gradually increase our believe about the parameter space with this particular choice of the prefactor. The dependence of the hyperparameter $\alpha$ on the number of observations guarantees that the precision of the Gaussian distributions increases with more observations, i.e. the standard deviation of the Gaussian decreases. This decrease in the standard deviation with the number of observations is necessary for the convergence of the approximative model to the objective function in the limit of an infinite number of observations.

		Several different protocols could be applied to increase the precision of the Gaussian distributions with the number of observations. In a scenario in which a dataset consisting of $n$ observations is approximated by a single Gaussian distribution, the precision of the Gaussian increases as the square root of the number of observed points, i.e. $\tau \propto \sqrt{n}$. Likewise we can consider a case in which we sample a random variable from the sum of Gaussian distributions. Assuming each of the Gaussian distributions has the same standard deviation $\sigma_0$, the standard deviation of the random variable is given by $\sqrt{n} \sigma_0$. 

		Based on these two considerations we studied the performance of \name with three different protocols for increasing the precision of the Gaussians with the number of observations: increasing the expected value of the precision with the number of observations as (i) $\langle\tau\rangle \propto n$, (ii) $\langle\tau\rangle \propto n^2$ and (iii) $\langle\tau\rangle \propto n^3$ given on the prior considerations. Results of the benchmark runs with \name and the three different schedules on selected objective functions are reported in Tab.~\ref{tab:shrinking_runs}. We report the minimum number of required objective function evaluations to reach values lower than the average lowest value encountered after $10^4$ random evaluations of the objective function (see Sec.~\ref{sec:random_search_results}). 

		\begin{table}[!ht]
			\setlength{\tabcolsep}{6pt}
			\centering
			\begin{tabular}{lcccc}
				\toprule
				Protocol & Ackley & Dejong & Schwefel & dAckley \\
				\midrule
				$\langle\tau\rangle \propto n$   & \error{39}{4} & \error{29}{2} & \error{137}{10} & \smerr{176}{8} \\
				$\langle\tau\rangle \propto n^2$ & \smerr{19}{1} & \smerr{21}{2} & \smerr{108}{10} & \error{179}{4} \\
				$\langle\tau\rangle \propto n^3$ & \error{43}{1} & \error{43}{1} & \error{126}{\phantom{1}6} & \error{186}{4} \\
				\bottomrule
			\end{tabular}
			\caption{Minimum number of required objective function evaluations to reach values lower than the average lowest value encountered after $10^4$ random evaluations of the objective function. For each simulation, the precision of the Gaussian distributions was increased with a different schedule based on the number of observations $n$. Lowest number of required evaluations for each objective function are printed in bold.}
			\label{tab:shrinking_runs}
		\end{table}

		We find, that the $\langle\tau\rangle \propto n^2$ schedule for shrinking approximating Gaussian distributions performs the best out of the proposed shrinking schedules across all objective functions. Slower increases in the precision of the Gaussian distributions like in the $\langle\tau\rangle \propto n$ schedule seem to create persistent regions in parameter space for which the acquisition function predicts unfavorable objective function values and does not sufficiently enhance exploration. In contrast, the $\langle\tau\rangle \propto n^3$ schedule shrinks Gaussian distributions too quickly such that acquired knowledge cannot be exploited sufficiently and the algorithm behaves more like random search. Based on these findings, \name adopts the $\langle\tau\rangle \propto n^2$ as shrinking schedule.

	\subsection{Benchmark results}\label{sec:benchmark_results}

		We benchmarked \namecomma RF and GP optimizers on a total of 15 different benchmark functions, which are reported and discussed in detail in Sec.~\ref{sec:supp_loss_functions}. For each of the benchmark function we ran the optimization algorithm for a total of $200$ iterations in $20$ independent runs initialized with different random seeds and recorded the lowest discovered objective function values for each iteration. Each objective function was optimized by three independent optimizer instances constructed from different exploration parameter values $\lambda \in \{ -1, 0, 1\}$. 

\begin{figure}[!ht]
			\centering
			\includegraphics[width = 1.0\textwidth]{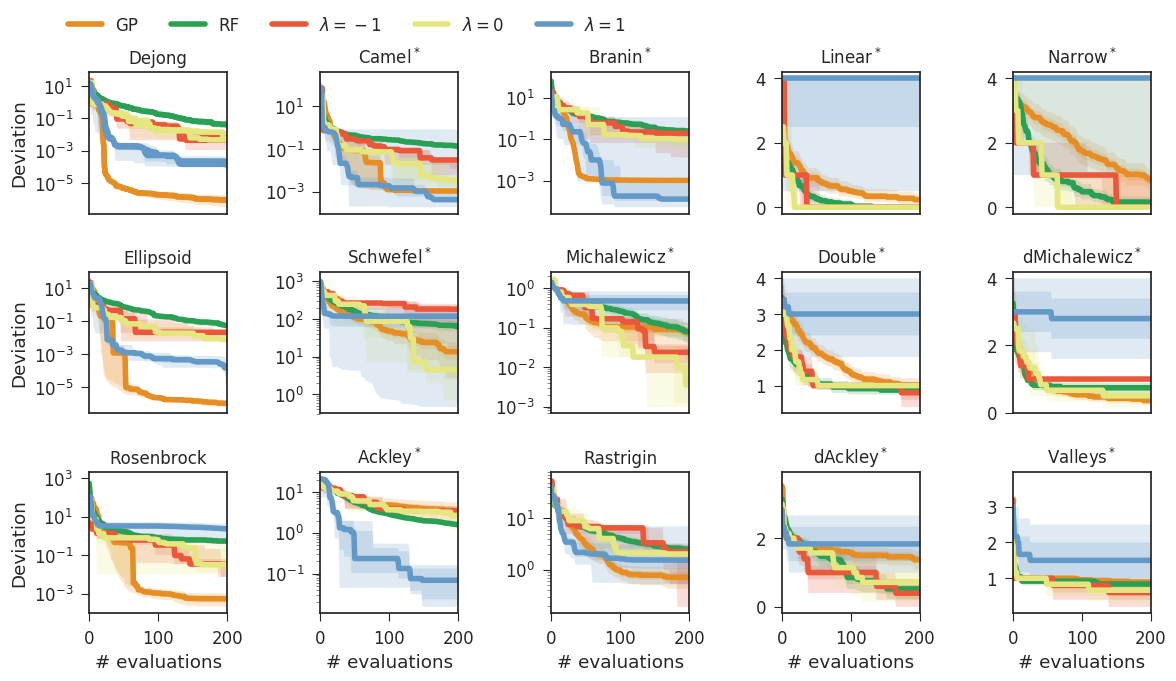}
			\caption{Lowest discovered objective function values averaged over $20$ independent optimizations for the optimization algorithm introduced in this study, which we constructed with three different choices for the exploration parameter $\lambda \in \{ -1, 0, 1\}$. We also report average lowest objective function values discovered by Bayesian optimization with Gaussian processes (spearmint package) and random forests (SMAC package). Objective functions for which the optimizer introduced in this study discovered the lowest objective function values are indicated by an asterisk. Uncertainty bands illustrate bootstrapped estimates of the deviation of the means with one and two standard deviations.}
			\label{fig:benchmark_results}
		\end{figure}

		Average lowest discovered objective function values and bootstrapped uncertainties are depicted in Fig.~\ref{fig:benchmark_results}. For comparison we provide the average lowest objective function values discovered by Bayesian optimization with Gaussian processes as implemented in spearmint and with random forests as implemented in SMAC.

		We find that instances of the optimization algorithm introduced in this study perform better than GP optimization and RF optimization in eleven out of 15 cases. For all studied objective function, instances of the introduced optimization algorithm perform better than RF optimization. GP optimization, however, discovers lower objective function values than instances of \name in four cases.

		These four cases, for which GP optimization discovers lower objective function values, are cases of purely convex objective functions (Dejong, Ellipsoid and Rosenbrock) and the Rastrigin function, which is mostly convex with small local minima modulating a general hyperparabolic shape. 

		We further observe that for some objective functions a more explorative (more negative) value of the exploration parameter improves the performance while in other cases a more exploitative (more positive) value is beneficial. For instance in the case of objective functions with a discrete co-domain (right and second to right columns in Fig.~\ref{fig:benchmark_results}) we observe good performance with more negative exploration parameter values. In fact, positive values for the exploration parameter result in the algorithm no longer finding the global minimum. This, however, can be explained by the fact that parameter points proposed with favoring exploitation will always be in close proximity to the current best observed parameter point. On a discrete co-domain, however, all points in close proximity to the current best yield the same objective function value.  

		In other cases, especially for mostly convex objective functions such as the Dejong function or the Camel function, \name performs the best when favoring exploitation over exploration. We therefore conclude that there cannot be a single best balance between exploration and exploitation for objective functions as diverse as the objective functions in this benchmark set.

	\subsection{Batch optimization results}\label{sec:swarm_exploration}

		Instead of marginalizing over the exploration parameter $\lambda$ in the acquisition function of \name, we suggest to propose parameter points in batches, some of which are determined favoring exploration and others favoring exploitation. Here, we demonstrate the the proposed algorithm benefits from such a procedure even if the proposed parameter points are evaluated sequentially. 

\begin{figure}[!ht]
			\centering
			\includegraphics[width = 1.0\textwidth]{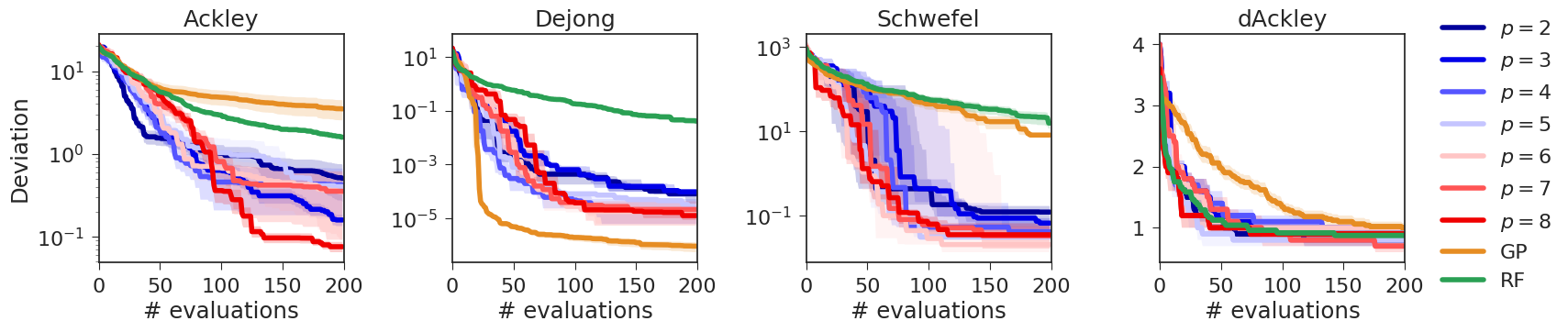}
			\caption{Lowest achieved average deviation between sampled objective function values and their global minima for a selected set of objective functions. \name proposed a total of $p$ points per batch, which was counted as $p$ function evaluations. Uncertainty bands illustrate bootstrapped estimates of the deviation of the means with one and two standard deviations.}
			\label{fig:swarm_exploration_results}
		\end{figure}

		Fig.~\ref{fig:swarm_exploration_results} displays the lowest achieved deviations between sampled objective function values and their global minima for a selected set of objective functions. Results reported for each objective function are averaged over $20$ independent runs. \name was run by proposing parameter points in batches of $p$ points, which where then evaluated sequentially. The evaluation of $p$ points in one batch was counted as $p$ objective function evaluations. We report the number of objective function evaluations needed to reach the indicated objective function values. Values for the exploration parameter were drawn evenly spaced from the $[-1, 1]$ interval. We provide the lowest achieved objective functions values in GP optimization and RF optimization for comparison.

		We observe that the synergistic effect of proposing parameter points in batches reported on the Ackley function in the main text (see Sec.~\ref{fig:swarm_behavior_one_loss}) also occurs for the other studied objective functions. In fact, even for objective functions with a discrete co-domain (right panel, Fig.~\ref{fig:swarm_exploration_results}) batch optimization seems to enhance the performance of \namecomma although we demonstrated that \name does not perform well on objective functions with a discrete co-domain when proposing parameter points with a bias towards exploitation.

	\subsection{Algorithm comparisons}

\begin{figure}[!ht]
			\centering
			\includegraphics[width = 1.0\textwidth]{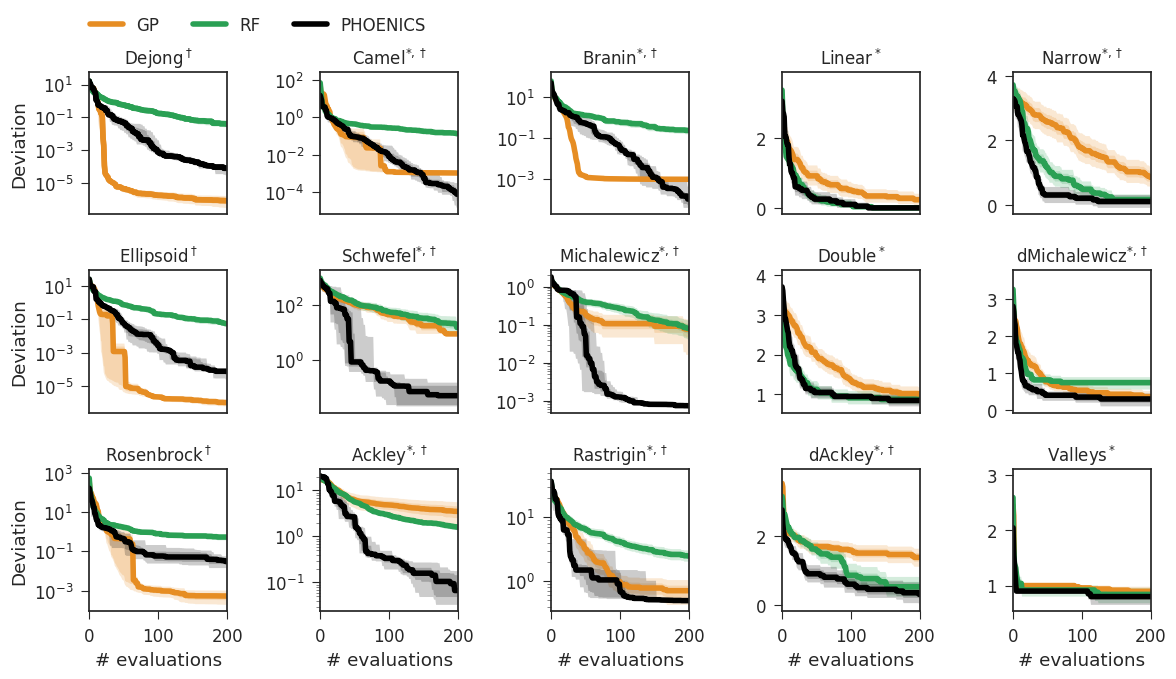}
			\caption{Deviations between lowest achieved objective function values and their global minima for a total of $20$ independent runs with the three studied optimization algorithms. \name was run in a batch exploration proposing $4$ points per batch based on exploration parameter values evenly spread across the $[-1, 1]$ interval. Objective functions on which \name performs better than GP and RF optimization are indicated denoted with $*$, function for which a synergistic effect was observed, i.e. improved performance with batch optimization compared to sequential optimization, are denoted with $\dagger$. Uncertainty bands illustrate bootstrapped estimates of the deviation of the means with one and two standard deviations.}
			\label{fig:all_comparisons}
		\end{figure}

		We demonstrated that \name generally performs better when proposing parameter points in batches, where some of the parameter points are proposed with a bias towards exploration and others are proposed with a bias towards exploitation (see Sec.~\ref{sec:batch_optimization}). Here, we report on the performance of \name with batch exploration and compare to RF and GP optimization. Fig.~\ref{fig:all_comparisons} depicts the traces of average deviations between the lowest achieved objective function values and their global minima for a total of $20$ independent runs with all three optimization algorithms. \name proposed $4$ points per batch based on exploration parameter values evenly spread across the $[-1, 1]$ interval. 

		We find that \name finds parameter points yielding objective function values closer to their global optima than RF and GP optimization in 12 out of 15 cases (indicated with $*$ in Fig.~\ref{fig:all_comparisons}). \name is only outperformed by GP optimization if the objective function is convex, i.e. for the Dejong, Ellipsoid and Rosenbrock function. Nevertheless, \name finds reasonable parameter points yielding low objective function values even for these functions. 

		In addition, we observe a synergistic effect of batch optimization for 12 out of 15 objective functions (indicated with $\dagger$ in Fig.~\ref{fig:all_comparisons}). For these objective functions, batch optimization performs better than any of the exploration parameter choices reported in Fig.~\ref{fig:benchmark_results}). We note, that the same batching protocol was used for all objective functions, indicating the flexibility and broad applicability of this protocol.

	\subsection{Belousov-Zhabotinsky reaction mechanism}\label{sec:belousov_zhabotinski}

		The Belousov-Zhabotinsky reaction is a prominent example of a nonlinear chemical oscillator.\cite{Zhabotinsky1967, Degn1967} While the detailed reaction mechanism is rather complex and involved a large number of elementary subreactions,\cite{Gyorgyi1990} the reaction can be summarized in three major subprocesses (see reaction~\ref{reaction:belousov-zhabotinsky}). \\

		\begin{Reaction*}
			\begin{align}
				\ce{
					2Br- + BrO3- + 3H+ + 3CH2(COOH)2 &-> 3BrCH(COOH)2 + 3H2O \tag{A}\label{reaction:A}\\
					BrO3- + 5H+ + 4Ce^{III} &-> 4Ce^{IV} + HOBr + 2 H2O \tag{B}\label{reaction:B}\\
					2Ce^{IV} + 2CH2(COOH)2 + BrCH(COOH)2 + HBrO + 2H2O &-> 2Ce^{III} + 2Br- + 3HOCH(COOH)2 + 4H+ \tag{C}\label{reaction:C}
				}
			\end{align}
			\caption{Subreactions of the Belousov-Zhabotinsky reaction.\cite{Kuhnert1987}}
			\label{reaction:belousov-zhabotinsky}
		\end{Reaction*}

		In reaction (\ref{reaction:A}) a bromite ion is reduced by a bromide ion through a series of two-electron reductions in which malonic acid reacts to bromomalonic acid. Reaction (\ref{reaction:B}) dominates over reaction (\ref{reaction:A}) at low bromide ion concentrations and forms $\text{Ce}^\text{IV}$ from $\text{Ce}^\text{III}$ while consuming bromite ions. Reaction (\ref{reaction:C}) then removes the $\text{Ce}^\text{IV}$ produced by reaction (\ref{reaction:B}).

		We model the dynamics of this reaction with the dimensionless Oregonator model (see Sec.~\ref{sec:applications} for details).\cite{Field1974} Numerical values for the parameter set used as a target for the optimization procedure are reported in Tab.~\ref{tab:ground_truth_reaction_constants}. We constrained the search space of parameter values based on the ranges reported in this table. Note, that particular choices of parameter sets within this bounded domain can result in quantitatively and qualitatively different dynamical behavior. In particular, parameter choices close to the target result in oscillatory behavior, for which the reduced concentrations $\alpha$, $\rho$ and $\eta$ change periodically over time, while other parameter choices can break the limit cycle and create a stable fixed point instead.\cite{Field1974, Gyorgyi1990, Voorsluijs2017} 

		\begin{table}[!ht]
			\setlength{\tabcolsep}{24pt}
			\centering
			\begin{tabular}{crc}
				\toprule 
					Parameter & Target & Range \\
				\midrule
					$s$ & $77.27$              & $0 \ldots 100$\\
					$w$ & $0.1610$             & $0 \ldots 1$\\
					$q$ & $8.375\cdot 10^{-6}$ & $10^{-8}\ldots 10^{-4}$ \\
					$f$ & $1$                  & $0 \ldots 5$\\
				\midrule
					$\alpha_0$ & $2.0\cdot 10^7$ & $10^4\ldots 10^9$\\
					$\eta_0$   & $3.3\cdot 10^3$ & $10^3\ldots 10^5$\\
					$\rho_0$   & $4.1\cdot 10^4$ & $10^3\ldots 10^6$\\
				\bottomrule
			\end{tabular}
			\caption{Reaction parameters of the reduced Oregonator model for the Belousov-Zhabotinsky reaction. Target parameters induce the existence of a limit cycle, from which chemical oscillations emerge. For finding these target parameters via optimization we constrained the domain space to the reported ranges. All reported quantities are dimensionless.}
			\label{tab:ground_truth_reaction_constants}
		\end{table}

\begin{figure}[!ht]
			\centering
			\includegraphics[width = 1.0\textwidth]{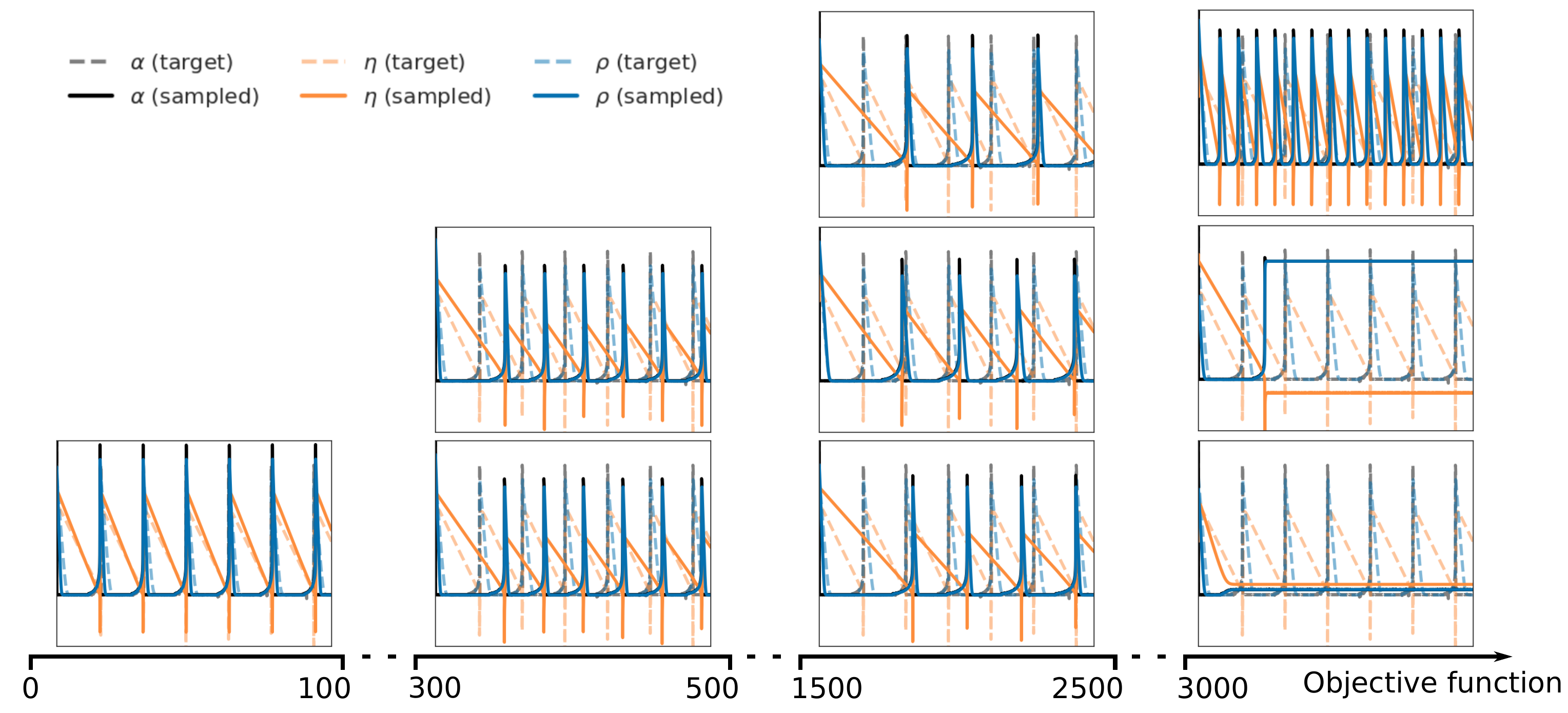}
			\caption{Examples for concentration traces related to different loss values. Concentration traces were obtained from different parameter sets all sampled by \name within a single optimization run. }
			\label{fig:oregonator_trace_plot_collection}
		\end{figure}

		Depending on the particular choice of reaction parameters for the Oregonator model the solutions of the differential equations can differ quantitatively and qualitatively. The set of target parameters features an oscillatory solution with a stable attractive limit cycle. Fig.~\ref{fig:oregonator_trace_plot_collection} illustrates different possible reduced concentration traces for different values of the constructed objective function. All presented concentration traces were sampled in a single optimization run of \name. 

		We find that loss values below $100$ closely resemble the target concentration traces. For such low loss values, we have qualitative and quantitative agreement between the traces. Loss values between $300$ and $500$ feature simulations for which the periodicity of the sampled concentration traces matches the periodicity of the target traces, but the traces are shifted by a phase. Slightly different periodicities are developed for losses between $1500$ and $2500$ while finally at losses above $3000$ the system shows rapid oscillations or even steady states.

		\putbib[main]
	\end{bibunit}

\end{document}